\documentclass[a4paper]{article}
\usepackage{MM805}
\usepackage{svg}
\usepackage{cite}
\usepackage{amsmath,amssymb,amsfonts, bbm}
\usepackage{subcaption}
\usepackage{graphicx}
\usepackage{textcomp}
\usepackage{multirow}
\usepackage{longtable}
\usepackage{multicol}
\usepackage{float}
\usepackage{url}
\usepackage{hyperref}
\usepackage{algorithm}
\usepackage{algorithmic}
\usepackage{multirow}

\usepackage{makecell}

\author{
  Tahmim Hossain \\
  University of Alberta \\
  \texttt{tahmim@ualberta.ca}
  \and
  Faisal Sayed \\
  University of Alberta\\
  \texttt{faisalz1@ualberta.ca}
  \and
  Solehin Islam \\
  University of Alberta \\
  \texttt{solehin1@ualberta.ca}
}
\title{Adaptive Local Binary Pattern: A Novel Feature Descriptor for Enhanced Analysis of Kidney Abnormalities in CT Scan Images using ensemble based Machine Learning Approach}

\date{}

\begin{document}
\maketitle
\section*{Abstract}

The shortage of nephrologists and the growing public health concern over renal failure have spurred the demand for AI systems capable of autonomously detecting kidney abnormalities. Renal failure, marked by a gradual decline in kidney function, can result from factors like cysts, stones, and tumors. Chronic kidney disease may go unnoticed initially, leading to untreated cases until they reach an advanced stage. The dataset, comprising 12,427 images from multiple hospitals in Dhaka, was categorized into four groups: cyst, tumor, stone, and normal. Our methodology aims to enhance CT scan image quality using Cropping, Resizing, and CALHE techniques, followed by feature extraction with our proposed Adaptive Local Binary Pattern (A-LBP) feature extraction method compared with the state-of-the-art local binary pattern (LBP) method. Our proposed features fed into classifiers such as Random Forest, Decision Tree, Naive Bayes, K-Nearest Neighbor, and SVM. We explored an ensemble model with soft voting to get a more robust model for our task. We got the highest of more than 99\% in accuracy using our feature descriptor and ensembling five classifiers (Random Forest, Decision Tree, Naive Bayes, K-Nearest Neighbor, Support Vector Machine) with the soft voting method.

\textit {Index Terms -} Feature Extraction, Kidney Disease, Classification, Adaptive Local Binary Pattern

\section{Introduction}

The global health community acknowledges CKD as a significant public health issue, emphasizing the importance of early detection, treatment, and its relationship with other chronic diseases for improving outcomes and implementing effective public health policies \cite{r1}. Chronic kidney disease poses a global health threat, with prevalence estimated at 8–16\%, driven primarily by diabetes mellitus, and associated with various complications including mortality, kidney disease progression, and cognitive decline, with the poorest populations facing the highest risk \cite{r2}. The forecasted rise of COPD, diabetes, chronic kidney disease, Alzheimer's disease, and lung cancer by 2040 suggests significant health system implications, with COPD projected to increase most notably from 9th to 4th place \cite{r3}. Kidney cancer ranks sixth among men in the US, with stones, cysts, and renal cell carcinoma being common kidney disorders affecting renal function. Approximately 6\% of women and 11\% of men in the US have experienced kidney stones \cite{r4}. Efficient identification of kidney-related issues is crucial due to their widespread occurrence and potential complications. Kidney stone disease affects 12\% of the global population, while renal cell carcinoma (RCC) is a prevalent malignancy. Kidney cysts, including simple and complex varieties, can impact renal and blood function, highlighting the need for accurate diagnosis to differentiate benign from malignant cases \cite{r5}. Timely identification of kidney abnormalities like cysts, stones, and tumors is essential for preventing renal failure, emphasizing the significance of early diagnosis in renal health management \cite{r6}. Rapid kidney-related radiological finding detection using artificial intelligence (AI) models has great potential to support medical staff and ease patient anxiety, solve the global radiology and nephrology shortage, and capitalize on advances in deep learning for visual tasks \cite{r5}.

\noindent The CT machine utilizes X-ray beams to scan specific areas of the body, generating cross-sectional images that offer detailed three-dimensional insights into the targeted anatomy, often employed alongside pathology tests for diagnosing kidney diseases \cite{r13}. Because CT scans may provide detailed slice-by-slice images as well as extensive three-dimensional data, they are especially useful when evaluating kidney function that's why it is better to use the CT scan image. Renal failure can be avoided by treating kidney abnormalities such as cysts, stones, and tumors as soon as possible \cite{r6}.

\noindent In this study, we introduce a novel feature descriptor tailored to improve the classification of medical CT scan images depicting kidney abnormalities. Our proposed descriptor, Adaptive Local Binary Pattern (ABPL), offers significant enhancements in discerning various kidney textures associated with normal, cystic, stony, and tumorous conditions. While conventional methods often rely on pre-trained models, CNNs, or established feature descriptors, our approach breaks new ground by developing a bespoke descriptor inspired by the Local Binary Pattern (LBP) technique. Through a thorough review of existing literature, we observed a prevalent trend towards utilizing pre-existing models or descriptors. However, our decision to innovate and craft a unique descriptor results in notable improvements across all evaluation metrics, marking a novel approach in this domain.

\noindent Our methodology involved several steps: firstly, we cropped the images to remove unwanted portions, then resized them to fit the model, and normalized the images. Subsequently, we employed the Contrast-Limited Adaptive Histogram Equalization (CLAHE) method to enhance image contrast while preserving details, particularly in areas with varying contrast levels, thus mitigating the risk of noise over-amplification in low-contrast regions. IBy reviewing different literature, we found several papers that utilized CLAHE as an image enhancement technique for their datasets. Building upon this, we implemented CLAHE as a preprocessing step in our methodology. Following preprocessing, we applied feature descriptors to extract relevant features from the images. Specifically, we utilized the Local Binary Pattern (LBP) as one feature descriptor and introduced a novel feature descriptor called Adaptive Local Binary Pattern (ALBP). These descriptors were crucial in capturing meaningful information from the images. Subsequently, we fed the extracted features into various classifiers, including Random Forest, Decision Tree, Naive Bayes, K-Nearest Neighbor, SVM, and a soft voting ensemble technique, for the classification of kidney abnormalities. Our experimental results revealed significant improvements in evaluation metrics, demonstrating the effectiveness of our proposed feature descriptor, as discussed in the experimental results section of this paper.

\section{Related Work}

 \noindent There are some recent research works where they tried to detect kidney abnormalities from the CT scan or MRI images by using different types of feature extraction techniques. Upon reviewing the extensive body of literature encompassed by papers authored by George et al. \cite{george2022kidney}, Georgieva et al. \cite{georgieva2016multistage}, Mamatov et al. \cite{mamatov20232d}, Xie et al. \cite{xie2020seru}, Vidyasaraswathi et al. \cite{vidyasaraswathi2015review}, and Abdoun \cite{abdoun2023using}, it becomes evident that the Contrast Limited Adaptive Histogram Equalization (CLAHE) method serves as a cornerstone in the enhancement of CT scan-based images. Its efficacy lies in its ability to accentuate details, mitigate noise, and enhance contrast, thereby facilitating improved visualization and interpretation of CT scans, which is crucial for accurate diagnosis and treatment planning in kidney-related pathologies.

 \noindent Hafizah et al. \cite{hafizah2012feature} opted for GLCM features, while Islam and Nahiduzzaman \cite{islam2022complex} leveraged CNNs for feature extraction, reflecting the growing utilization of deep learning in this domain. The utilization of Darknet-19 features by Baygin et al. \cite{baygin2022exemplar} underscores the diversity of approaches, as does the combination of CRF with HOG by Bhole et al. \cite{bhole20113d}. Meanwhile, Vasanthselvakumar et al. \cite{vasanthselvakumar2019detection} focused exclusively on HOG, and Christiyana and Rajamani \cite{christiyana2014new} proposed a fusion of Laws Masks and Local Binary Patterns, showcasing the varied strategies employed to extract meaningful features from kidney images. Furthermore, Ali \cite{ali2017studying} and Kang et al. \cite{kang2022renal} explored Local Binary Patterns and 3D CNN with ConvLSTM, respectively, highlighting the versatility in feature extraction methodologies adopted across the literature.

\noindent In classification techniques, authors have made strategic choices tailored to the complexities of kidney image analysis. Chunduru et al. \cite{chunduru2024multi} selected CNNs and Random Forests as classifiers, a decision likely influenced by the robustness of these methods in handling high-dimensional data and complex decision boundaries. Meanwhile, Gharaibeh et al. \cite{gharaibeh2022radiology} conducted a thorough review of various Deep Learning models alongside traditional machine learning approaches, providing valuable insights into the comparative efficacy of different methodologies. Moreover, Xv et al. \cite{xv2021machine} introduced a support vector machine (SVM) coupled with feature selection algorithms, offering a nuanced approach to feature engineering and model optimization. Conversely, Qadir et al. \cite{qadir2023kidney} explored a hybrid model integrating DenseNet201 and Random Forest, underscoring the trend towards hybrid architectures that hold the strengths of multiple methodologies. Similarly, Chak et al. \cite{chak2020neural} endeavored to develop a kidney image classifier using a combination of SVM and NN, showcasing the synergy between traditional machine learning and deep learning approaches. On the application side, Soni and Rai \cite{soni2020kidney} focused on kidney stone detection, employing Histogram Equalization and embossing techniques in conjunction with SVM as the classifier. This highlights the importance of methodologies to specific clinical challenges within the broader domain of kidney imaging. In a similar, Alshebly et al. \cite{alshebly2019prediction} utilized neural networks and Logistic Regression for predicting chronic kidney disease, illustrating the diverse array of predictive modeling techniques employed in the literature. The prevalence of random forest classifiers in papers by Amiri et al. \cite{amiri2021radiomics} and Alex et al. \cite{alex2022hybrid} underscores their utility in handling complex datasets characteristic of kidney imaging. Moreover, Alex et al. \cite{alex2022hybrid} introduced a modified hybrid random forest classifier specifically for predicting kidney damage, demonstrating an approach to algorithmic development.

\noindent Furthermore, Islam et al. \cite{islam2022vision} employed VGG16 and Swin Transformer for detecting kidney stones, cysts, and tumors, showcasing the integration of state-of-the-art deep learning architectures for comprehensive organ-level analysis. Similarly, Pande et al. \cite{r5} implemented YoLo v8 for predicting kidney abnormalities, highlighting the potential of advanced object detection techniques in this domain. Finally, across papers by Reuben et al. \cite{reuben2024effective}, Vasanthselvakumar et al. \cite{vasanthselvakumar2020automatic}, Alzu’bi et al. \cite{alzu2022kidney}, Parakh et al. \cite{parakh2019urinary}, Kuo et al. \cite{kuo2019automation}, Chaki and Ucar \cite{chaki2024efficient}, Tushar et al. \cite{tushar2021classification}, and Majid et al. \cite{majid2023enhanced}, deep learning methods emerge as the predominant approach for addressing kidney abnormality issues. This underscores the growing trend towards leveraging deep learning's capacity for representation learning and feature abstraction in medical imaging tasks, particularly in the context of kidney disease diagnosis and prognosis.

\section{Proposed Feature Extraction Method}
In this section, the proposed feature extraction method is presented. The aim is to capture not only the background texture but also the inherent features within CT scan images. Drawing inspiration from the Local Binary Pattern (LBP) approach, a novel method termed Adaptive Local Binary Pattern (A-LBP) is introduced. This section explains the details of the proposed feature descriptor, the A-LBP method.

\begin{figure} [H]
    \centering 
	\includegraphics[width=0.75\textwidth]{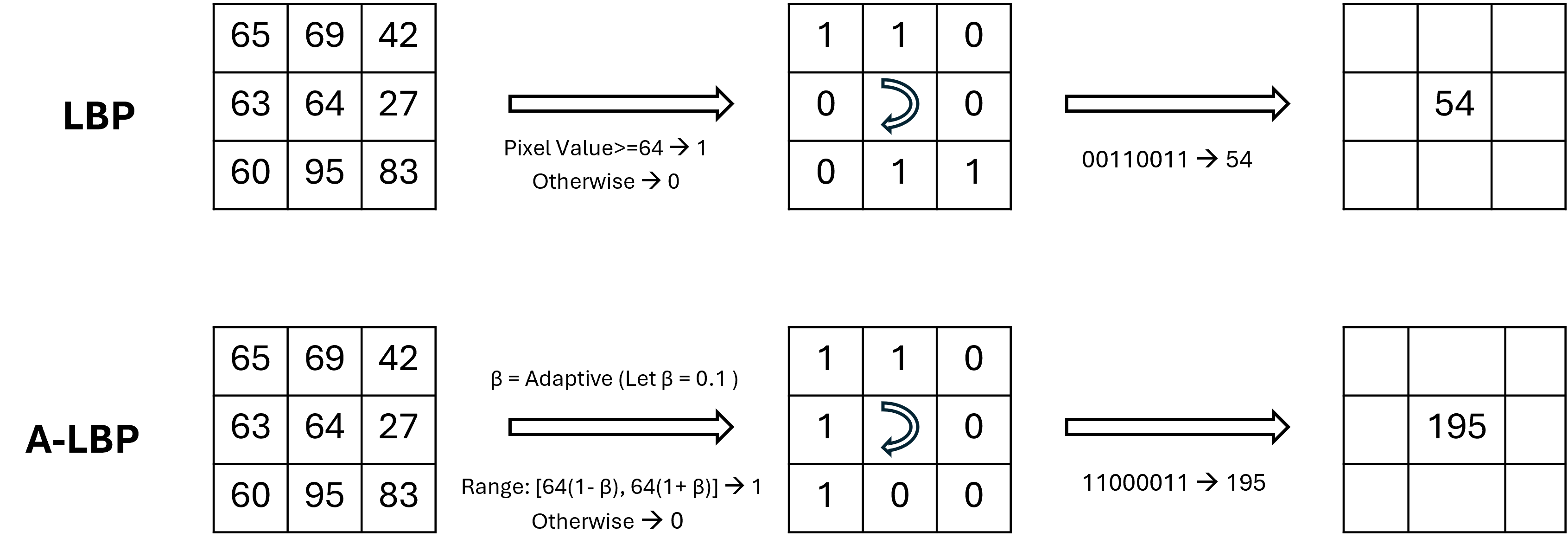}
	\caption{The encoding processes of LBP and A-LBP in a pixel block.}
	\label{fig::1}
\end{figure}

\begin{algorithm}[H]
\caption{Proposed Adaptive Binary Pattern (A-LBP) Algorithm}
\label{alg:1}

\textbf{Input:} Input image (matrix), Center pixel coordinates $(x_{c}, y_{c})$ and Adaptive parameter $\beta$ 

\textbf{Output:} Binary pattern corresponding to the center pixel and its decimal value

\begin{algorithmic}[1]

\STATE Let $I_c(x_c, y_c)$ represent the center pixel value of the image at coordinates $(x_c, y_c)$ and neighborhood pixel value as $I_n$ and Assigned value as $V_n$
\STATE Define upper and lower bounds as $UBV$ \& $LBV$ where, \\  $UBV = I_c \times (1 + \beta)$ snd $LBV = I_c\times (1 - \beta)$
\FOR{each pixel in the 8-neighborhood of $(x_c, y_c)$}
    \IF{$LBV \leq I_n \leq UBV$}
        \STATE Set $I_n$ = 1 = $V_n$
    \ELSE
        \STATE Set $I_n$ = 0 = $V_n$
    \ENDIF
\ENDFOR
\STATE Construct a binary pattern by traversing the neighborhood pixel values in a clockwise direction, starting from the upper left as LSB to MSB.
\STATE Convert the binary pattern into a decimal value using $\sum_{n=0}^{N-1} V_n \times 2^{n}$ equation.
\STATE Replace center pixel value $C(x, y)$ with acquired decimal value from the binary pattern
\end{algorithmic}
\end{algorithm}

\noindent The first process block in figure \ref{fig::1} demonstrates the operation of LBP. It involves taking the center value of the pixel and comparing it with the values of the 8 neighboring pixels. If a neighbor pixel value is greater than or equal to the center pixel value, it is replaced with 1; otherwise, it is replaced with 0. Next, a binary pattern is constructed, starting with the top-left value as the least significant bit (LSB) and proceeding clockwise to the value as the most significant bit (MSB). The binary pattern is then converted into a decimal value, which is subsequently utilized to replace the center pixel value. To understand the second process block in figure \ref{fig::1}, the reference to our proposed algorithm \ref{alg:1} provides further clarity.

\noindent To explain the second process block from the figure \ref{fig::1}, we can check out the equation \ref{eq:1}. 
\begin{equation}
    F_{\text{texture}} = \text{ALBP}_{((x_{c}, y_{c}), \beta)} = \sum_{n=0}^{N-1} V_n \times 2^{n}
    \label{eq:1}
\end{equation}

\noindent In equation \ref{eq:1}, $V_n$ denotes the value assigned to the $n^{th}$ neighboring pixel based on a thresholding function. Then, the ALBP  value is the decimal value of the binary pattern. In ${ALBP}_{((x_{c}, y_{c}), \beta)}$ for the center pixel at coordinates $(x_c,y_c)$ and the adaptive value $\beta$ is computed as a weighted sum of the binary values assigned to its neighboring pixels. To determine the threshold function, let $I(x_c,y_c)$ denote the intensity of the center pixel at coordinates $(x_c,y_c)$ in the input image. The ALBP value for this pixel is computed based on the comparison of its intensity with the intensities of its neighboring pixels within a defined threshold range.The thresholding function, denoted as $T(I_c,I_n, \beta)$ , compares the intensity of the center pixel $I_c$ with the intensity of each neighboring pixel $I_n$. The condition we put as the threshold range is given in equations \ref{eq:2}.

\begin{equation}
    V_n =\begin{cases} (1-\beta)\times{I_c} \leq {I_n} \leq (1+\beta)\times{I_c} & V_n = 1\\otherwise &  V_n = 0\end{cases}  
    \label{eq:2}
\end{equation}

\section{Dataset}
\noindent 12,446 unique PACS images from different hospitals in Dhaka, Bangladesh were included in the dataset \cite{r7}. The photos showed patients with kidney tumors, cysts, normal conditions, or kidney stones. From contrast and non-contrast investigations of the whole abdomen and urogram protocols, both coronal and axial cuts were included. The DICOM data have been converted into lossless JPEG. The author of the dataset mentioned that to guarantee data accuracy, each image finding was verified by a medical technician. 3,709 photographs of cysts, 5,077 images of normal circumstances, 1,377 images of kidney stones, and 2,283 images of tumors.

\section{Proposed Framework}

\begin{figure} [H]
    \centering 
	\includegraphics[width=0.8\textwidth]{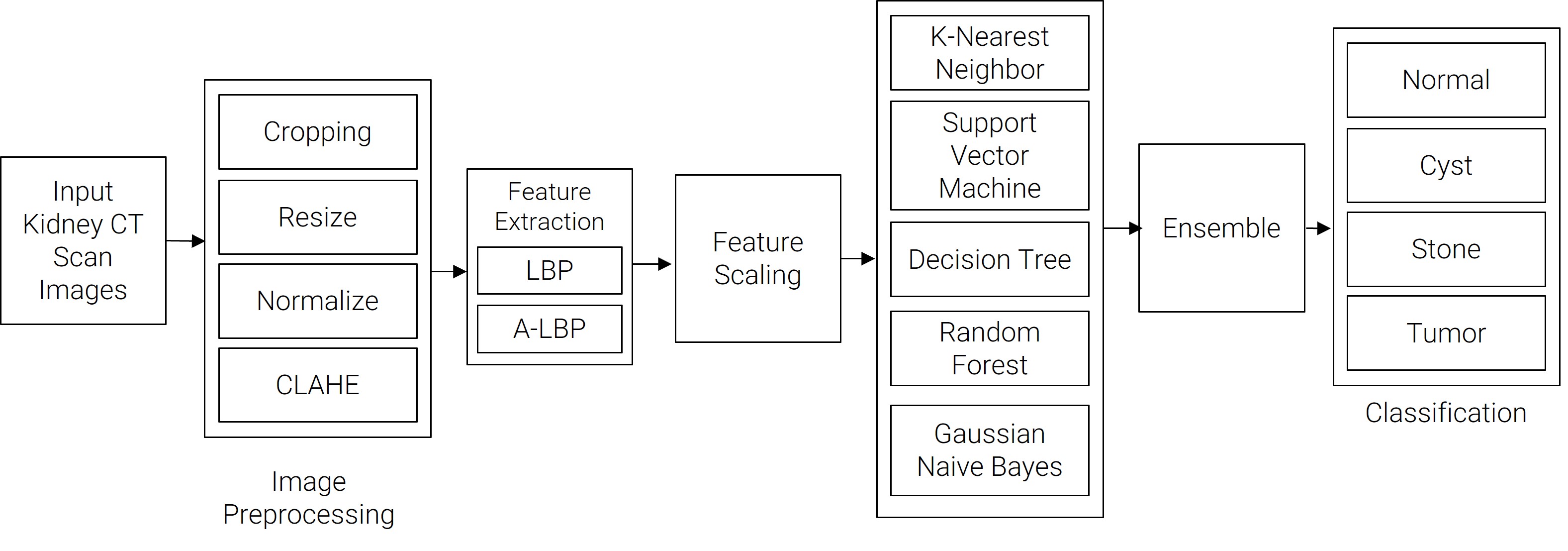}
	\caption{Proposed Methodology}
	\label{fig::0}
\end{figure}

\noindent In this section, the proposed framework has been discussed. Figure \ref{fig::0} is the proposed architecture of our work.

\subsection{Data Preprocessing}

\subsubsection{Data Cropping \& Resizing}
\noindent The initial phase of our image processing workflow involves meticulously cropping the CT scan images to isolate and emphasize the pertinent anatomical structures. This cropping process serves a dual purpose: first, it effectively eliminates extraneous background noise and non-essential elements from the images, thus enhancing the clarity and focus on the regions of interest. Second, it ensures that subsequent analyses are conducted with precision and efficiency, as they are directed toward the critical anatomical features essential for diagnosis and evaluation. Figure \ref{fig::2}(a) is a raw dataset after that it has been cropped as shown in figure \ref{fig::2}(b). 

\noindent Following the cropping stage, we proceed to standardize the resolution of the images to a uniform size, typically resizing them into a standardized dimension of 224x224 pixels. This standardization is crucial for several reasons. Firstly, it facilitates consistency across the dataset, ensuring that all images are represented in a comparable format, which is essential for reliable and accurate analysis. Standardizing the resolution also helps to streamline subsequent computational processes, as it reduces variability and simplifies the integration of various image processing techniques and algorithms. Figure \ref{fig::2} (c) is an image after resizing it into 224x224 pixels.

\subsubsection {Apply Contrast-Limited Adaptive Histogram Equalization (CLAHE) to the Data}

\begin{figure} [H]
    \centering 
	\includegraphics[width=0.8\textwidth]{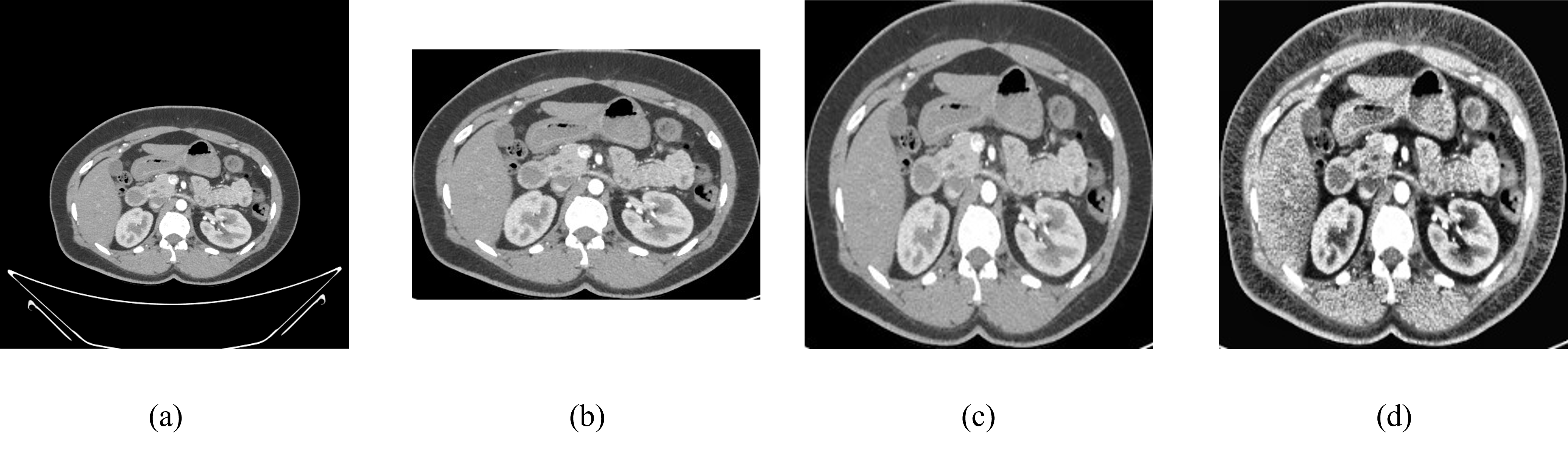}
	\caption{(a) Raw Normal Kidney Image (b) Cropped Image (c) Resized Image (d) CLAHE Applied Image}
	\label{fig::2}
\end{figure}

\noindent CALHE, or Contrast Adaptive Local Histogram Equalization, is an innovative technique that enhances image contrast by adjusting to local intensity variations, unlike traditional global histogram equalization methods. By considering the local context, CALHE proves particularly valuable for identifying subtle anatomical structures and anomalies in medical images. Its adaptive approach revolutionizes visualization, offering unprecedented precision and detail, thus advancing both research and clinical practice in radiology and beyond. Figure \ref{fig::2} (d) is an image after applying CLAHE which enhances the image more than regular raw data.

\subsection{Feature Extraction}
For feature extraction, we used one of the handcrafted texture analysis methods named Local Binary Pattern compared with our proposed Feature Descriptor named Adaptive Local Binary Pattern. 

\subsubsection{Local Binary Pattern (LBP)}
One promising method in medical image analysis is the categorization of renal illness using Local Binary Patterns (LBP). A texture descriptor called LBP compares a pixel's intensity to that of its neighboring pixels to extract characteristics from a picture. LBP can identify textural patterns that indicate pathological changes in renal tissue, which helps distinguish between healthy and sick regions when it comes to the classification of kidney disease. Accurate and automated diagnosis of kidney disorders can be accomplished by integrating LBP features into classification algorithms, such as support vector machines or neural networks. This allows for prompt interventions and customized treatment plans. The types of textural features that are obtained after applying a Local Binary Pattern are illustrated in Figure \ref{fig::3}.

\subsubsection{Our Proposed Feature Extraction Method, Adaptive Local Binary Pattern (A-LBP)}

\begin{figure} [H]
    \centering 
	\includegraphics[width=0.5\textwidth]{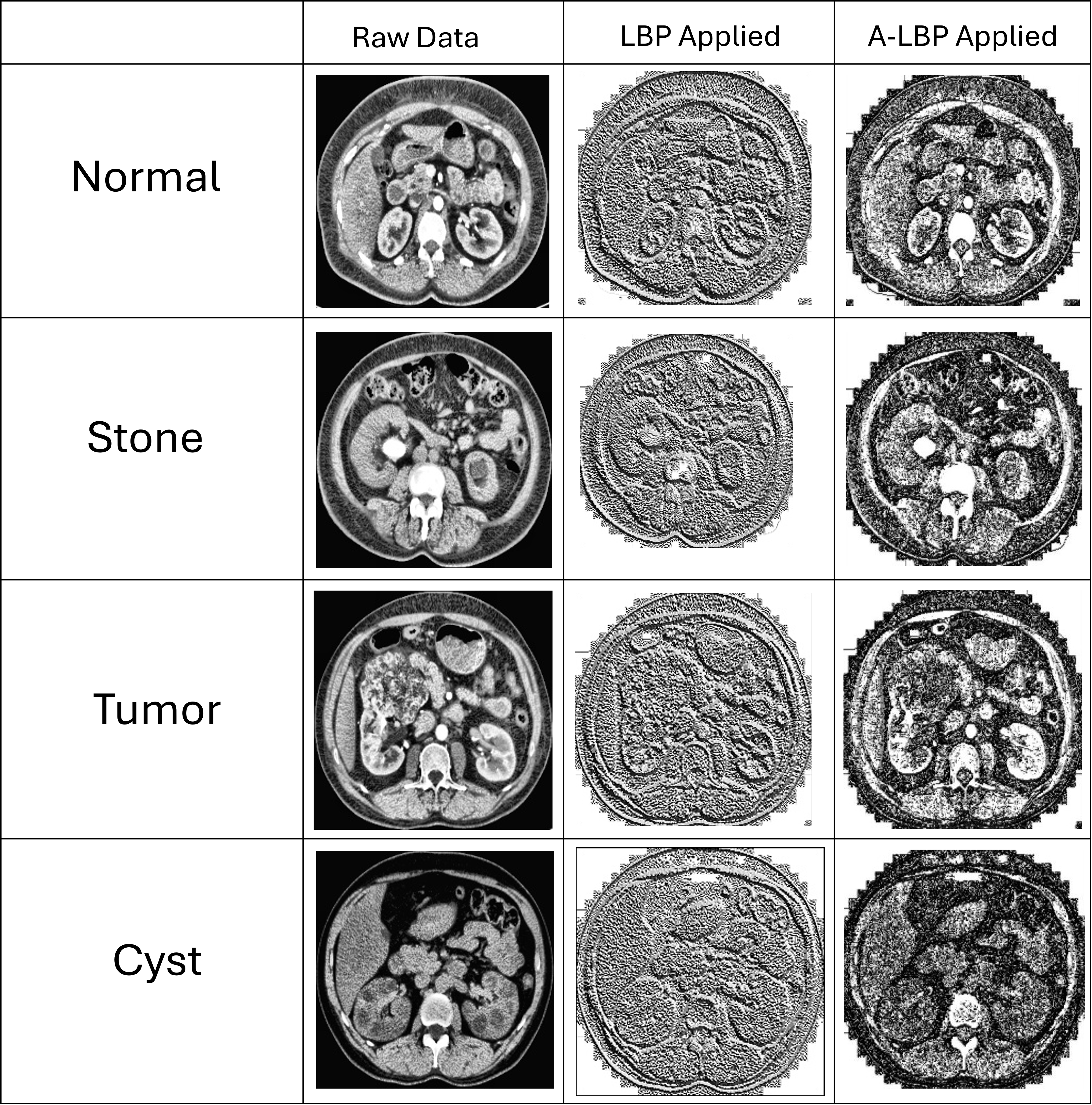}
	\caption{Four types of kidney image slices after applying Local Binary Pattern (LBP) and our proposed feature descriptor, Adaptive Local Binary Pattern(A-LBP)}
	\label{fig::3}
\end{figure}

We present Adaptive Local Binary Pattern (A-LBP), a novel feature extraction technique that aims to extract intrinsic features from CT scan images as well as background texture. A-LBP provides a more sophisticated method that takes local context into account, and it takes inspiration from Local Binary Pattern (LBP). The algorithm and accompanying figure describe the A-LBP method, which compares pixel intensities with neighboring pixels within a threshold range. Each pixel block's descriptor is the binary pattern that is produced after being converted to a decimal value. A-LBP is a descriptor that captures both structural and textural information. Through the use of A-LBP, physicians are able to obtain a better understanding of the subtle differences found in kidney images, which is important for correctly classifying healthy and diseased tissue.  Specifically, A-LBP captures nuanced texture patterns indicative of renal pathologies, thereby facilitating automated diagnosis and personalized treatment strategies. The features that are obtained after applying the Adaptive Local Binary Pattern are illustrated in Figure \ref{fig::3} comparing with the Local Binary Pattern.

\subsection{Classifier Selection} 
\noindent The given features were utilized to train the classifiers listed below, and a soft voting ensemble technique was proposed to improve the robustness of the decision-making process for detecting the four types of kidney diseases. 

\subsubsection{Random Forest}
\noindent As part of its ensemble learning process, Random Forest builds a large number of decision trees during training and outputs a class that represents the mean prediction (regression) or mode of the classes (classification) of each individual tree. Because each tree in the random forest is constructed using a random subset of the dataset's features, overfitting is minimized and robustness is raised. Random Forest gives more consistent and accurate results by aggregating the predictions of several decision trees.

\subsubsection{Decision Tree}
\noindent A supervised learning technique called decision trees is applied to tasks involving regression and classification. In order to maximize the homogeneity of the target variable within each subset, it recursively partitions the dataset into subsets based on the value of a chosen feature. Decision trees are widely used because of their interpretability and simplicity in understanding, as well as their capacity to capture intricate decision boundaries.

\subsubsection{Naive Bayes}
\noindent The "naive" assumption of feature independence underlies the probabilistic classification technique known as Naive Bayes, which is based on the Bayes theorem. Naive Bayes is a simple algorithm that works well in practice, especially when it comes to text categorization jobs. Given a set of features, it determines the likelihood of each class and uses the class with the highest probability as the forecast.

\subsubsection{K-Nearest Neighbor (K-NN)}
\noindent Finding the majority class among a given data point's k nearest neighbors is how the straightforward and user-friendly K-Nearest Neighbor classification technique operates. The class with the highest frequency among the neighbors is allocated to the data point, and the choice of k sets the number of neighbors considered. K-NN mostly depends on the distance metric that's used to gauge how similar two data points are to one another.

\subsubsection{Support Vector Machine (SVM)}
\noindent Provide assistance For classification and regression applications, Support Vector Machine is a potent supervised learning algorithm. In order to divide the data points of various classes with the largest margin possible, it finds the ideal hyperplane. Because SVMs may handle nonlinear decision boundaries by utilizing a variety of kernel functions, they are both effective and versatile in high-dimensional spaces.

\subsubsection{Soft Voting Ensemble}
By averaging the probability assigned to each class, the Soft Voting Ensemble is a meta-estimator classifier that aggregates the predictions of several basic classifiers. It enables varied contributions from each classifier according to its degree of confidence, leading to a more sophisticated decision-making process. By utilizing the advantages of a variety of base classifiers, a soft voting ensemble can enhance the classification task's overall performance and robustness.

\section{Experimental Results}
In this section, the experimental results that we have obtained from the proposed project have been discussed.
\subsection{Experimental Settings}
The dataset was divided into 80\% for training and 20\% for testing across all cases. Our proposed feature extraction method (Adaptive Local Binary Pattern (A-LBP) ), along with Local Binary Pattern (LBP), was employed on the preprocessed dataset. The classifiers mentioned in section 5.3 were utilized. Evaluation metrics including precision, recall, and F1 Score were employed to assess and compare the effectiveness of the models. No GPU was utilized for training; instead, an Intel Xeon 2.20 GHz CPU with 32GB of RAM was employed. Training and evaluation processes were conducted on Kaggle.

\subsection{For Detecting the Normal Kidney Images}

In order to identify kidney abnormalities of type Cyst in CT scan images, the table compares the performance of two feature descriptors for classification: Local Binary Pattern (LBP) and Adaptive Local Binary Pattern (ALBP). In terms of test accuracy, precision, recall, and F1 score, ALBP consistently performs better than LBP across a range of classifiers. For example, Random Forest obtained a test accuracy of 91\% using the LBP descriptor, whereas ALBP greatly increased the accuracy to 97\%. In a similar vein, ALBP outperformed LBP in terms of precision, recall, and F1 scores for every classifier. Interestingly, ALBP scored almost perfectly on most metrics, demonstrating its excellence in identifying the complex traits associated with anomalies of the cystic kidney. On the other hand, the performance of LBP varied greatly among classifiers, with certain ones obtaining comparatively less precision and accuracy. 

\begin{table}[H]
\centering
\resizebox{\columnwidth}{!}{%
\begin{tabular}{|cccccc|}
\hline
\multicolumn{6}{|c|}{\textbf{Kidney Type: Normal}}                                                                                                                    \\ \hline
\multicolumn{1}{|c|}{\textbf{\begin{tabular}[c]{@{}c@{}}Feature \\ Descriptor\end{tabular}}} &
  \multicolumn{1}{c|}{\textbf{Classifier}} &
  \multicolumn{1}{c|}{\textbf{\begin{tabular}[c]{@{}c@{}}Test \\ Accuracy\end{tabular}}} &
  \multicolumn{1}{c|}{\textbf{Precision}} &
  \multicolumn{1}{c|}{\textbf{Recall}} &
  \textbf{F1 Score} \\ \hline
\multicolumn{1}{|c|}{\multirow{6}{*}{\begin{tabular}[c]{@{}c@{}}Local Binary\\  Pattern\\ (LBP)\end{tabular}}} &
  \multicolumn{1}{c|}{Random Forest (RF)} &
  \multicolumn{1}{c|}{.91} &
  \multicolumn{1}{c|}{.88} &
  \multicolumn{1}{c|}{.99} &
  .93 \\ \cline{2-6} 
\multicolumn{1}{|c|}{} & \multicolumn{1}{c|}{Decision Tree (DT)}            & \multicolumn{1}{c|}{.80} & \multicolumn{1}{c|}{.83}  & \multicolumn{1}{c|}{.85}  & .84  \\ \cline{2-6} 
\multicolumn{1}{|c|}{} & \multicolumn{1}{c|}{Naive Bayes (NB)}              & \multicolumn{1}{c|}{.53} & \multicolumn{1}{c|}{.66}  & \multicolumn{1}{c|}{.47}  & .55  \\ \cline{2-6} 
\multicolumn{1}{|c|}{} & \multicolumn{1}{c|}{K-Nearest Neighbor (K-NN)}     & \multicolumn{1}{c|}{.96} & \multicolumn{1}{c|}{.98}  & \multicolumn{1}{c|}{.96}  & .97  \\ \cline{2-6} 
\multicolumn{1}{|c|}{} & \multicolumn{1}{c|}{Support Vector Machine (SVM)} & \multicolumn{1}{c|}{.89} & \multicolumn{1}{c|}{.85}  & \multicolumn{1}{c|}{1.00}  & .92  \\ \cline{2-6} 
\multicolumn{1}{|c|}{} & \multicolumn{1}{c|}{Soft Voting Ensemble}          & \multicolumn{1}{c|}{.88} & \multicolumn{1}{c|}{.88}  & \multicolumn{1}{c|}{.95}  & .92  \\ \hline
\multicolumn{1}{|c|}{\multirow{6}{*}{\begin{tabular}[c]{@{}c@{}}Adaptive Local \\ Binary Pattern\\  (ALBP)\end{tabular}}} &
  \multicolumn{1}{c|}{Random Forest (RF)} &
  \multicolumn{1}{c|}{.97} &
  \multicolumn{1}{c|}{.95} &
  \multicolumn{1}{c|}{1.00} &
  .98 \\ \cline{2-6} 
\multicolumn{1}{|c|}{} & \multicolumn{1}{c|}{Decision Tree (DT)}            & \multicolumn{1}{c|}{.80} & \multicolumn{1}{c|}{.85}  & \multicolumn{1}{c|}{.87}  & .86  \\ \cline{2-6} 
\multicolumn{1}{|c|}{} & \multicolumn{1}{c|}{Naive Bayes (NB)}              & \multicolumn{1}{c|}{.68} & \multicolumn{1}{c|}{.76}  & \multicolumn{1}{c|}{.81}  & .78  \\ \cline{2-6} 
\multicolumn{1}{|c|}{} & \multicolumn{1}{c|}{K-Nearest Neighbor (K-NN)}     & \multicolumn{1}{c|}{.99} & \multicolumn{1}{c|}{1.00} & \multicolumn{1}{c|}{1.00} & 1.00 \\ \cline{2-6} 
\multicolumn{1}{|c|}{} & \multicolumn{1}{c|}{Support Vector Machine (SVM)} & \multicolumn{1}{c|}{.99} & \multicolumn{1}{c|}{1.00} & \multicolumn{1}{c|}{1.00} & 1.00 \\ \cline{2-6} 
\multicolumn{1}{|c|}{} & \multicolumn{1}{c|}{Soft Voting Ensemble}          & \multicolumn{1}{c|}{.99} & \multicolumn{1}{c|}{.99}  & \multicolumn{1}{c|}{1.00} & 1.00 \\ \hline
\end{tabular}
}
\caption{Kidney Type: Normal}
\label{Normal}
\end{table}

\subsection{For Detecting the Stone in Kidney Images}
The performance of classifiers based on the feature descriptor is highlighted in the analysis of table \ref{Stone}, with an emphasis on finding the best descriptor to recognize kidney stones. K-Nearest Neighbor (K-NN) is the best classifier using the Local Binary Pattern (LBP) descriptor; it achieves a remarkable 96\% test accuracy and shows strong precision, recall, and an F1 score of 92\%. On the other hand, Naive Bayes (NB) shows a consistent lower score on all criteria, suggesting that it is not very effective for this task. But the descriptor that really stands out is the Adaptive Local Binary Pattern (ALBP) one. Classifiers that use ALBP, like K-NN, SVM, and the Soft Voting Ensemble, routinely achieve remarkable accuracy rates of 99\% along with perfect precision, recall, and F1 scores. Significantly, Naive Bayes consistently demonstrates poor performance, confirming its inadequacy for precise kidney stone categorization. Therefore, while both descriptors have their strengths, the ALBP descriptor emerges as the superior choice for effectively identifying kidney stones, as evidenced by the outstanding performance of classifiers utilizing this feature descriptor.

\begin{table}[H]
\centering
\resizebox{\columnwidth}{!}{%
\begin{tabular}{|cccccc|}
\hline
\multicolumn{6}{|c|}{\textbf{Kidney Type: Stone}}                                                                                                                     \\ \hline
\multicolumn{1}{|c|}{\textbf{\begin{tabular}[c]{@{}c@{}}Feature \\ Descriptor\end{tabular}}} &
  \multicolumn{1}{c|}{\textbf{Classifier}} &
  \multicolumn{1}{c|}{\textbf{\begin{tabular}[c]{@{}c@{}}Test \\ Accuracy\end{tabular}}} &
  \multicolumn{1}{c|}{\textbf{Precision}} &
  \multicolumn{1}{c|}{\textbf{Recall}} &
  \textbf{F1 Score} \\ \hline
\multicolumn{1}{|c|}{\multirow{6}{*}{\begin{tabular}[c]{@{}c@{}}Local Binary\\  Pattern\\ (LBP)\end{tabular}}} &
  \multicolumn{1}{c|}{Random Forest (RF)} &
  \multicolumn{1}{c|}{.91} &
  \multicolumn{1}{c|}{.98} &
  \multicolumn{1}{c|}{.69} &
  .81 \\ \cline{2-6} 
\multicolumn{1}{|c|}{} & \multicolumn{1}{c|}{Decision Tree (DT)}            & \multicolumn{1}{c|}{.80} & \multicolumn{1}{c|}{.62}  & \multicolumn{1}{c|}{.61}  & .62  \\ \cline{2-6} 
\multicolumn{1}{|c|}{} & \multicolumn{1}{c|}{Naive Bayes (NB)}              & \multicolumn{1}{c|}{.53} & \multicolumn{1}{c|}{.13}  & \multicolumn{1}{c|}{.27}  & .18  \\ \cline{2-6} 
\multicolumn{1}{|c|}{} & \multicolumn{1}{c|}{K-Nearest Neighbor (K-NN)}     & \multicolumn{1}{c|}{.96} & \multicolumn{1}{c|}{.92}  & \multicolumn{1}{c|}{.92}  & .92  \\ \cline{2-6} 
\multicolumn{1}{|c|}{} & \multicolumn{1}{c|}{Support Vector Machine (SVM):} & \multicolumn{1}{c|}{.89} & \multicolumn{1}{c|}{1.00}  & \multicolumn{1}{c|}{.56}  & .72  \\ \cline{2-6} 
\multicolumn{1}{|c|}{} & \multicolumn{1}{c|}{Soft Voting Ensemble}          & \multicolumn{1}{c|}{.88} & \multicolumn{1}{c|}{.86}  & \multicolumn{1}{c|}{.69}  & .76  \\ \hline
\multicolumn{1}{|c|}{\multirow{6}{*}{\begin{tabular}[c]{@{}c@{}}Adaptive Local \\ Binary Pattern\\  (ALBP)\end{tabular}}} &
  \multicolumn{1}{c|}{Random Forest (RF)} &
  \multicolumn{1}{c|}{.97} &
  \multicolumn{1}{c|}{1.00} &
  \multicolumn{1}{c|}{.84} &
  .92 \\ \cline{2-6} 
\multicolumn{1}{|c|}{} & \multicolumn{1}{c|}{Decision Tree (DT)}            & \multicolumn{1}{c|}{.80} & \multicolumn{1}{c|}{.59}  & \multicolumn{1}{c|}{.61}  & .60  \\ \cline{2-6} 
\multicolumn{1}{|c|}{} & \multicolumn{1}{c|}{Naive Bayes (NB)}              & \multicolumn{1}{c|}{.68} & \multicolumn{1}{c|}{.39}  & \multicolumn{1}{c|}{.67}  & .49  \\ \cline{2-6} 
\multicolumn{1}{|c|}{} & \multicolumn{1}{c|}{K-Nearest Neighbor (K-NN)}     & \multicolumn{1}{c|}{.99} & \multicolumn{1}{c|}{1.00} & \multicolumn{1}{c|}{1.00} & 1.00 \\ \cline{2-6} 
\multicolumn{1}{|c|}{} & \multicolumn{1}{c|}{Support Vector Machine (SVM)} & \multicolumn{1}{c|}{.99} & \multicolumn{1}{c|}{1.00} & \multicolumn{1}{c|}{.98}  & .99  \\ \cline{2-6} 
\multicolumn{1}{|c|}{} & \multicolumn{1}{c|}{Soft Voting Ensemble}          & \multicolumn{1}{c|}{.99} & \multicolumn{1}{c|}{1.00} & \multicolumn{1}{c|}{.97}  & .98  \\ \hline
\end{tabular}
}
\caption{Kidney Type: Stone}
\label{Stone}
\end{table}

\subsection{For Detecting the Tumor in Kidney Images}
ALBP-based classifiers routinely achieve superior performance compared to LBP-based classifiers across multiple parameters in Table \ref{Tumor}. For instance, ALBP-driven Random Forest (RF) classifiers exhibit a remarkable test accuracy of 97\%, coupled with a precision of 100\%, recall of 97\%, and an F1 score of 98\%. In contrast, LBP-based RF classifiers achieve a lower accuracy of 91\%, along with a precision of 95\%, recall of 79\%, and an F1 score of 86\%. Similarly, ALBP-based classifiers such as K-Nearest Neighbor (K-NN), Support Vector Machine (SVM), and the Soft Voting Ensemble consistently demonstrate higher accuracy rates and F1 scores compared to their LBP counterparts. This consistent trend underscores the superiority of ALBP over LBP in accurately identifying kidney tumors, making it a compelling choice for medical imaging tasks requiring precise classification.

\begin{table}[H]
\centering
\resizebox{\columnwidth}{!}{%
\begin{tabular}{|cccccc|}
\hline
\multicolumn{6}{|c|}{\textbf{Kidney Type: Tumor}}                                                                                                                     \\ \hline
\multicolumn{1}{|c|}{\textbf{\begin{tabular}[c]{@{}c@{}}Feature \\ Descriptor\end{tabular}}} &
  \multicolumn{1}{c|}{\textbf{Classifier}} &
  \multicolumn{1}{c|}{\textbf{\begin{tabular}[c]{@{}c@{}}Test \\ Accuracy\end{tabular}}} &
  \multicolumn{1}{c|}{\textbf{Precision}} &
  \multicolumn{1}{c|}{\textbf{Recall}} &
  \textbf{F1 Score} \\ \hline
\multicolumn{1}{|c|}{\multirow{6}{*}{\begin{tabular}[c]{@{}c@{}}Local Binary\\  Pattern\\ (LBP)\end{tabular}}} &
  \multicolumn{1}{c|}{Random Forest (RF)} &
  \multicolumn{1}{c|}{.91} &
  \multicolumn{1}{c|}{.95} &
  \multicolumn{1}{c|}{.79} &
  .86 \\ \cline{2-6} 
\multicolumn{1}{|c|}{} & \multicolumn{1}{c|}{Decision Tree (DT)}            & \multicolumn{1}{c|}{.80} & \multicolumn{1}{c|}{.71}  & \multicolumn{1}{c|}{.68}  & .69  \\ \cline{2-6} 
\multicolumn{1}{|c|}{} & \multicolumn{1}{c|}{Naive Bayes (NB)}              & \multicolumn{1}{c|}{.53} & \multicolumn{1}{c|}{.47}  & \multicolumn{1}{c|}{.47}  & .47  \\ \cline{2-6} 
\multicolumn{1}{|c|}{} & \multicolumn{1}{c|}{K-Nearest Neighbor (K-NN)}     & \multicolumn{1}{c|}{.96} & \multicolumn{1}{c|}{.93}  & \multicolumn{1}{c|}{.94}  & .93  \\ \cline{2-6} 
\multicolumn{1}{|c|}{} & \multicolumn{1}{c|}{Support Vector Machine (SVM)} & \multicolumn{1}{c|}{.89} & \multicolumn{1}{c|}{.96}  & \multicolumn{1}{c|}{.77}  & .85  \\ \cline{2-6} 
\multicolumn{1}{|c|}{} & \multicolumn{1}{c|}{Soft Voting Ensemble}          & \multicolumn{1}{c|}{.88} & \multicolumn{1}{c|}{.91}  & \multicolumn{1}{c|}{.71}  & .80  \\ \hline
\multicolumn{1}{|c|}{\multirow{6}{*}{\begin{tabular}[c]{@{}c@{}}Adaptive Local \\ Binary Pattern\\  (ALBP)\end{tabular}}} &
  \multicolumn{1}{c|}{Random Forest (RF)} &
  \multicolumn{1}{c|}{.97} &
  \multicolumn{1}{c|}{1.00} &
  \multicolumn{1}{c|}{.97} &
  .98 \\ \cline{2-6} 
\multicolumn{1}{|c|}{} & \multicolumn{1}{c|}{Decision Tree (DT)}            & \multicolumn{1}{c|}{.80} & \multicolumn{1}{c|}{.75}  & \multicolumn{1}{c|}{.75}  & .75  \\ \cline{2-6} 
\multicolumn{1}{|c|}{} & \multicolumn{1}{c|}{Naive Bayes (NB)}              & \multicolumn{1}{c|}{.68} & \multicolumn{1}{c|}{.53}  & \multicolumn{1}{c|}{.50}  & .51  \\ \cline{2-6} 
\multicolumn{1}{|c|}{} & \multicolumn{1}{c|}{K-Nearest Neighbor (K-NN)}     & \multicolumn{1}{c|}{.99} & \multicolumn{1}{c|}{1.00} & \multicolumn{1}{c|}{1.00} & 1.00 \\ \cline{2-6} 
\multicolumn{1}{|c|}{} & \multicolumn{1}{c|}{Support Vector Machine (SVM)} & \multicolumn{1}{c|}{.99} & \multicolumn{1}{c|}{1.00} & \multicolumn{1}{c|}{1.00} & 1.00 \\ \cline{2-6} 
\multicolumn{1}{|c|}{} & \multicolumn{1}{c|}{Soft Voting Ensemble}          & \multicolumn{1}{c|}{.99} & \multicolumn{1}{c|}{1.00} & \multicolumn{1}{c|}{.97}  & .98  \\ \hline
\end{tabular}
}
\caption{Kidney Type: Tumor}
\label{Tumor}
\end{table}

\subsection{For Detecting the Cyst in Kidney Images}
The table \ref{Cyst} compares the performance of different classifiers using Local Binary Pattern (LBP) and Adaptive Local Binary Pattern (ALBP) feature descriptors to detect cysts from kidney. A-LBP Performance was outstanding. Using ALBP, the classifiers achieved high accuracy, with Random Forest and Support Vector Machine (SVM) reaching 99\% accuracy and a perfect 1.00 in precision, recall, and F1 score. Both LBP and ALBP benefited from the soft voting ensemble technique, showing improved robustness in classification but with ALPB the results were remarkable and far better than what we had with LPB.
\begin{table}[H]
\centering
\resizebox{\columnwidth}{!}{%
\begin{tabular}{|cccccc|}
\hline
\multicolumn{6}{|c|}{\textbf{Kidney Type: Cyst}}                                                                                                                      \\ \hline
\multicolumn{1}{|c|}{\textbf{\begin{tabular}[c]{@{}c@{}}Feature \\ Descriptor\end{tabular}}} &
  \multicolumn{1}{c|}{\textbf{Classifier}} &
  \multicolumn{1}{c|}{\textbf{\begin{tabular}[c]{@{}c@{}}Test \\ Accuracy\end{tabular}}} &
  \multicolumn{1}{c|}{\textbf{Precision}} &
  \multicolumn{1}{c|}{\textbf{Recall}} &
  \textbf{F1 Score} \\ \hline
\multicolumn{1}{|c|}{\multirow{6}{*}{\begin{tabular}[c]{@{}c@{}}Local Binary\\  Pattern\\ (LBP)\end{tabular}}} &
  \multicolumn{1}{c|}{Random Forest (RF)} &
  \multicolumn{1}{c|}{.91} &
  \multicolumn{1}{c|}{.93} &
  \multicolumn{1}{c|}{.97} &
  .95 \\ \cline{2-6} 
\multicolumn{1}{|c|}{} & \multicolumn{1}{c|}{Decision Tree (DT)}            & \multicolumn{1}{c|}{.80} & \multicolumn{1}{c|}{.89}  & \multicolumn{1}{c|}{.88}  & .89  \\ \cline{2-6} 
\multicolumn{1}{|c|}{} & \multicolumn{1}{c|}{Naive Bayes (NB)}              & \multicolumn{1}{c|}{.53} & \multicolumn{1}{c|}{.73}  & \multicolumn{1}{c|}{.73}  & .73  \\ \cline{2-6} 
\multicolumn{1}{|c|}{} & \multicolumn{1}{c|}{K-Nearest Neighbor (K-NN)}     & \multicolumn{1}{c|}{.96} & \multicolumn{1}{c|}{.96}  & \multicolumn{1}{c|}{.97}  & .96  \\ \cline{2-6} 
\multicolumn{1}{|c|}{} & \multicolumn{1}{c|}{Support Vector Machine (SVM):} & \multicolumn{1}{c|}{.89} & \multicolumn{1}{c|}{.93}  & \multicolumn{1}{c|}{.96}  & .85  \\ \cline{2-6} 
\multicolumn{1}{|c|}{} & \multicolumn{1}{c|}{Soft Voting Ensemble}          & \multicolumn{1}{c|}{.88} & \multicolumn{1}{c|}{.88}  & \multicolumn{1}{c|}{.96}  & .92  \\ \hline
\multicolumn{1}{|c|}{\multirow{6}{*}{\begin{tabular}[c]{@{}c@{}}Adaptive Local \\ Binary Pattern\\  (ALBP)\end{tabular}}} &
  \multicolumn{1}{c|}{Random Forest (RF)} &
  \multicolumn{1}{c|}{.97} &
  \multicolumn{1}{c|}{.98} &
  \multicolumn{1}{c|}{.99} &
  .99 \\ \cline{2-6} 
\multicolumn{1}{|c|}{} & \multicolumn{1}{c|}{Decision Tree (DT)}            & \multicolumn{1}{c|}{.80} & \multicolumn{1}{c|}{.85}  & \multicolumn{1}{c|}{.82}  & .84  \\ \cline{2-6} 
\multicolumn{1}{|c|}{} & \multicolumn{1}{c|}{Naive Bayes (NB)}              & \multicolumn{1}{c|}{.68} & \multicolumn{1}{c|}{.88}  & \multicolumn{1}{c|}{.60}  & .71  \\ \cline{2-6} 
\multicolumn{1}{|c|}{} & \multicolumn{1}{c|}{K-Nearest Neighbor (K-NN)}     & \multicolumn{1}{c|}{.99} & \multicolumn{1}{c|}{1.00} & \multicolumn{1}{c|}{1.00} & 1.00 \\ \cline{2-6} 
\multicolumn{1}{|c|}{} & \multicolumn{1}{c|}{Support Vector Machine (SVM):} & \multicolumn{1}{c|}{.99} & \multicolumn{1}{c|}{.99}  & \multicolumn{1}{c|}{1.00} & 1.00 \\ \cline{2-6} 
\multicolumn{1}{|c|}{} & \multicolumn{1}{c|}{Soft Voting Ensemble}          & \multicolumn{1}{c|}{.99} & \multicolumn{1}{c|}{1.00} & \multicolumn{1}{c|}{1.00} & 1.00 \\ \hline
\end{tabular}
}
\caption{Kidney Type: Cyst}
\label{Cyst}
\end{table}

\section{Discussion}
Table \ref{Results} showcases a comparative analysis between existing studies on kidney disease diagnosis and our research. Our model has made significant advancements in the field, surpassing the performance of previous models and making notable contributions to the domain. Our approach, which uses an ensemble of five models (Random Forest, Decision Tree, Naive Bayes, KNN, and SVM) along with Adaptive Local Binary Pattern (A-LBP), achieves an impressive accuracy of 0.99 when compared to existing kidney stone detection methods. This accuracy level matches, if not exceeds, several of the well-known techniques covered in the table \ref{Results}. Unlike some other approaches that concentrate on certain classifiers or algorithms, such as CNN, our strategy makes use of the synergy between numerous models, utilizing their strengths to improve overall performance. Furthermore, by using A-LBP, our approach shows flexibility to local characteristics, which is important for correctly detecting kidney abnormalities in a variety of complicated datasets. Overall, the kidney stone detection capabilities of our suggested technique are promising and have the potential to be robust in medical imaging.

\begin{table}[H]
\centering
\resizebox{\columnwidth}{!}{%
\begin{tabular}{|c|c|c|c|c|}
\hline
\textbf{Authors} &
  \textbf{Method} &
  \textbf{Algorithm/Classifier Used} &
  \textbf{Dataset} &
  \textbf{Accuracy} \\ \hline
\makecell{Mehmet et al.\cite{baygin2022exemplar}} &
  Exemplar Darknet &
  KNN &
  \begin{tabular}[c]{@{}c@{}}1799 CT images \\ - 790 kidney stones \\ - 1009 healthy classes\end{tabular} &
  0.99 \\ \hline
\makecell{Vasanthselvakumar R et al.\cite{vasanthselvakumar2020automatic}} &
  HoG, CNN &
  Adaboost Classifier &
  \begin{tabular}[c]{@{}c@{}}90 ultrasound images\\ - 91 kidney stones\\ - 92 Cystic Kidney\\ - 87 Kidney Cancer\end{tabular} &
  \begin{tabular}[c]{@{}c@{}}CNN - 0.85\\ SIFT with SVM - 0.82\\ SURF with SVM - 0.84\end{tabular} \\ \hline
\begin{tabular}[c]{@{}c@{}}
\makecell{Chelladurai CALLINS \\ CHRISTIYANA et al.\cite{christiyana2014new}}\end{tabular} &
  LFLBP &
  - &
  \begin{tabular}[c]{@{}c@{}}365 images\\  -Normal, \\ - Cortical Cysts\\  - Medical Renal Diseases\end{tabular} &
  \begin{tabular}[c]{@{}c@{}} Normal – 0.81\\ CC – 0.63\\ MRD – 0.86\end{tabular} \\ \hline
\makecell{Jyoti Verma et al.\cite{verma2017analysis}} &
  PCA &
  KNN, SVM &
  \begin{tabular}[c]{@{}c@{}}- Abnormal Kidney\\ -Normal Kidney\end{tabular} &
  \begin{tabular}[c]{@{}c@{}}KNN - 0.89\\ SVM - 0.84\end{tabular} \\ \hline
\makecell{Anilkumar Chunduru\cite{chunduru2024multi}} &
  CNN &
  \begin{tabular}[c]{@{}c@{}}Logistic Regression,\\ XGBoost,\\ KNN,\\ Navie Bayes,\\ SVM\end{tabular} &
  \begin{tabular}[c]{@{}c@{}}- 3889 infected Kidney\\ - 1349 Normal\end{tabular} &
  \begin{tabular}[c]{@{}c@{}}Logistic Regression – 0.76,\\ XGBoost –  0.72,\\ KNN - 0.70,\\ Navie Bayes- 0.71,\\ SVM - 0.75\end{tabular} \\ \hline
\makecell{Kadir et.al\cite{yildirim2021deep}} &
  - &
  XResNet-50 &
  \begin{tabular}[c]{@{}c@{}}1799 images\\ - Normal\\ - Stone\end{tabular} &
  0.96 \\ \hline
\makecell{Abdalbasit Mohammed Qadir\cite{qadir2023kidney}} &
  DenseNet201 &
  Random Forest &
  \begin{tabular}[c]{@{}c@{}}5400 CT images\\ - 1350 Cystic Kidney\\ - 1350 Normal Kidney\\ - 1350 Stone Kidney\\ - 1350 Tumor Kidney\end{tabular} &
  0.99 \\ \hline
\makecell{Sudharson et.al\cite{sudharson2020ensemble}} &
  \begin{tabular}[c]{@{}c@{}}ResNet-101, ShuffleNet, \\ and MobileNet-v2\end{tabular} &
  SVM &
  \begin{tabular}[c]{@{}c@{}}4420 CT images\\ - 1105 Cystic Kidney\\ - 1105 Normal Kidney\\ - 1105 Stone Kidney\\ - 1105 Tumor Kidney\end{tabular} &
  0.96 \\ \hline
\makecell{Dalia et.al\cite{alzu2022kidney}} &
  Canny Edge Detection &
  \begin{tabular}[c]{@{}c@{}}VGG16,\\ ResNet50,\\ 2D CNN\end{tabular} &
  \begin{tabular}[c]{@{}c@{}}8600 CT images\\ - Normal\\ - Tumor\end{tabular} &
  \begin{tabular}[c]{@{}c@{}}VGG16- 0.60\\ ResNet50 - 0.96,\\ 2D CNN – 0.97\end{tabular} \\ \hline
\makecell{Md Nazmul et.al\cite{islam2022vision}} &
  - &
  \begin{tabular}[c]{@{}c@{}}EANET, \\ Swin Transformers, \\ CCT, \\ VGG16, \\ Inception V3, \\ Resnet50.\end{tabular} &
  \begin{tabular}[c]{@{}c@{}}12446 CT images\\ - 3709 Cystic Kidney\\ - 5077 Normal Kidney\\ - 1377 Stone Kidney\\ - 2283 Tumor Kidney\end{tabular} &
  \begin{tabular}[c]{@{}c@{}}EANET - 0.77, \\ Swin Transformers - 0.99, \\ CCT - 0.96, \\ VGG16 - 0.98, \\ Inception V3 - 0.61, \\ Resnet50 - 0.73.\end{tabular} \\ \hline
\textbf{Our Proposed Method} &
  \textbf{\begin{tabular}[c]{@{}c@{}}Adaptive Local Binary Pattern \\ (A-LBP)\end{tabular}} &
  \textbf{\begin{tabular}[c]{@{}c@{}}Ensemble five models\\  (Random Forest,\\ Decision Tree,\\ Naïve Bayes,\\ KNN,\\ SVM)\end{tabular}} &
  \textbf{\begin{tabular}[c]{@{}c@{}}12446 CT images\\ - 3709 Cystic Kidney\\ - 5077 Normal Kidney\\ - 1377 Stone Kidney\\ - 2283 Tumor Kidney\end{tabular}} &
  \textbf{0.99} \\ \hline
\end{tabular}%
}
\caption{Performance metrics comparison of other models and our proposed method.}
\label{Results}
\end{table}

\section{Novelty}
Several innovative contributions to the categorization of kidney anomalies in CT scan pictures are presented by this effort. The main contribution is the creation of the Adaptive Local Binary Pattern (ALBP) feature descriptor, which is designed to distinguish between different kidney textures. In contrast to traditional techniques, our method innovates by creating a custom description, leading to significant gains in all assessment criteria. Furthermore, new uses of picture preprocessing methods like Contrast-Limited Adaptive Histogram Equalization (CLAHE) improve our classification accuracy even further. By implementing these advances, we are able to identify kidney problems in CT scan pictures with greater accuracy. Also to run this program a person doesn't need a GPU which is environment-friendly and it can run on a regular machine.

\section{Conclusion \& Future Work}
The Adaptive Local Binary Pattern (ALBP) feature descriptor improves the classification of kidney anomalies in medical CT scan pictures, according to the study's conclusion. A unique custom descriptor was created to distinguish across different kidney textures, and this led to notable gains in all assessment criteria. It will make it easier for the medical professional to identify kidney abnormalities. This study emphasizes how crucial innovative feature extraction techniques are to raising the accuracy of medical picture analysis. Upgrading preprocessing methods, investigating sophisticated machine learning algorithms, and verifying results on larger datasets could be the main goals of future work to raise the bar for kidney anomaly categorization.

\bibliographystyle{plain}
\bibliography{Project_Report}
\end{document}